\documentclass[runningheads]{llncs}
\usepackage{graphicx}
\usepackage{amsmath,amssymb} 
\usepackage{color}
\usepackage[width=122mm,left=12mm,paperwidth=146mm,height=193mm,top=12mm,paperheight=217mm]{geometry}

\usepackage{bm}
\usepackage{subfigure}
\usepackage[linesnumbered,ruled]{algorithm2e}

\def\ie{{\emph{i.e.}}}
\def\eg{{\emph{e.g.}}}
\def\etal{{\emph{et al.~}}}

\def\x{{\mathbf x}}
\def\w{{\mathbf w}}
\def\X{{\mathbf X}}
\def\y{{\mathbf y}}

\def\e{{\mathbf e}}

\def\D{{\mathbf D}}
\def\k{{\mathbf k}}
\def\K{{\mathbf K}}
\def\f{{\mathbf f}}
\def\q{{\mathbf q}}
\def\F{{\mathcal F}}
\def\aa{{\bm \alpha}}

\begin{document}
\pagestyle{headings}
\mainmatter

\title{Real-Time Visual Tracking: Promoting the Robustness of Correlation Filter Learning} 

\titlerunning{Real-Time Visual Tracking: Promoting the Robustness of Correlation Filter Learning}

\authorrunning{Yao Sui, Ziming Zhang, Guanghui Wang, Yafei Tang, Li Zhang}

\author{Yao Sui$^1$, Ziming Zhang$^2$, Guanghui Wang$^1$, Yafei Tang$^3$, Li Zhang$^4$}


\institute{$^1$Dept. of EECS, University of Kansas, Lawrence, KS 66045, USA\\$^2$Dept. of ECE, Boston University, Boston, MA 02215, USA\\$^3$China Unicom Research Institute, Beijing 100032, China\\$^4$Dept. of EE, Tsinghua University, Beijing 100084, China\\
\email{suiyao@gmail.com, zzhang14@bu.edu, ghwang@ku.edu, tangyf24@chinaunicom.cn, chinazhangli@tsinghua.edu.cn}
}

\maketitle

\begin{abstract}
  Correlation filtering based tracking model has received lots of attention and achieved great success in real-time tracking, however, the lost function in current correlation filtering paradigm could not reliably response to the appearance changes caused by occlusion and illumination variations. This study intends to promote the robustness of the correlation filter learning. By exploiting the anisotropy of the filter response, three sparsity related loss functions are proposed to alleviate the overfitting issue of previous methods and improve the overall tracking performance. As a result, three real-time trackers are implemented. Extensive experiments in various challenging situations demonstrate that the robustness of the learned correlation filter has been greatly improved via the designed loss functions. In addition, the study reveals, from an experimental perspective, how different loss functions essentially influence the tracking performance. An important conclusion is that the sensitivity of the peak values of the filter in successive frames is consistent with the tracking performance. This is a useful reference criterion in designing a robust correlation filter for visual tracking.
\end{abstract}

\section{Introduction}
In recent years, there is a significant interest in correlation filtering based tracking. Under this paradigm, a correlation filter is efficiently learned online from previously obtained target regions, and the target is located according to the magnitude of the filter response over a large number of target candidates. The main strength of this paradigm is its high computational efficiency, because the target and the candidate regions can be represented in frequency domain and manipulated by fast Fourier transform (FFT), which yields $\mathcal{O}\left(n\log n\right)$ computational complexity for a region of $\sqrt{n}\times\sqrt{n}$ pixels. For this reason, extensive real-time trackers \cite{Bolme2010,Henriques2012,Danelljan2014a,Danelljan2014b,Li2014a,Zhang2014c,Henriques2015,Liu2015,Danelljan2015} have been proposed within the correlation filtering paradigm.

Specifically, a correlation filter is learned from previously obtained target regions to approximate an expected filter response, such that the peak of the response is located at the center of the target region. The response used in previous methods is often assigned to be of Gaussian shaped, which is treated as a continuous version of an impulse signal. For this reason, the learned filter is encouraged to produce Gaussian shaped response. The candidate region with the strongest filter response is determined as the target.

\begin{figure}[t]
  \centering
  \subfigure[]{
  \includegraphics[width=0.14\linewidth]{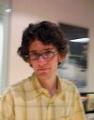}
  }
  \subfigure[]{
  \label{fig:samples}
  \includegraphics[width=0.14\linewidth]{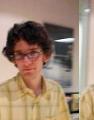}
  \includegraphics[width=0.14\linewidth]{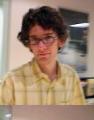}
  \includegraphics[width=0.14\linewidth]{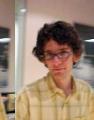}
  \includegraphics[width=0.14\linewidth]{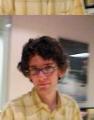}
  }
  \caption{Illustration of the cyclic shift method. (a) The base image. (b) The cyclic shifts of the base image with $\pm15$ pixels in horizontal and vertical directions, respectively.}
  \label{fig:cyclic}
\end{figure}
Note that the Gaussian shaped response, from a signal processing perspective, is \emph{isotropic}, \ie, all the regions that deviate the same distance away from the center of the target are assigned to the same response values. However, it has been demonstrated that the anisotropic response values can significantly improve the tracking performance from a regression point of view \cite{Shunli2015,Hare2011}\footnote{The exact equivalence between regression and correlation filtering is proved in \cite{Henriques2015}.}, \eg, using the overlap rates between the training image samples and the target as the response values.  Fig. \ref{fig:cyclic} illustrates a popular approach for samples generation adopted by previous correlation filtering based trackers \cite{Henriques2012,Henriques2015}. It is evident from Fig. \ref{fig:samples} that the regions of interest are not continuous. This will bring challenges to the correlation filter learning if the response values of the four significantly different regions are enforced to be the same, easily leading to an overfitting.

In addition, from a loss function point of view, the correlation filter is always learned under the squared loss (\ie, $\ell_2$-loss) in the previous methods. The choice for the squared loss is limited by the Parseval's identity, by which the learning problem can be exactly transferred into frequency domain. Moreover, the squared loss can lead to a closed-form solution, which guarantees the high computational efficiency. Nevertheless, the target appearance may change significantly during tracking in various challenging situations, such as occlusion and illumination variation. A robust loss function is required to reliably respond to these appearance changes, and avoid the overfitting. The squared loss allows the filter response to fit the expected response with small errors, \ie, stochastically yields Gaussian errors with a small variance. In the presence of significant appearance changes, the errors may be extremely large in some feature dimensions, leading to an instability of the squared loss.

Inspired by the previous success, an \emph{anisotropy} of the filter response is exploited in this work by means of an adaptive learning approach via robust loss functions, including $\ell_1$-, $\ell_1\ell_2$-, and $\ell_{2,1}$-loss functions. These loss functions will increase the robustness, since they allow large errors in the filter learning in the presence of significant appearance changes. As a result, three real-time trackers are proposed in this study, and it is demonstrated how the loss functions essentially influence the tracking performance. An interesting observation is obtained from the experimental results, which can be taken as a reference criterion in designing a robust correlation filter for visual tracking: the sensitivity of the peak values of the filter in successive frames is consistent with the tracking performance. The proposed algorithms are evaluated by extensive experiments on a popular benchmark \cite{Wu2013}, and they outperform the competing counterparts.

\section{Related Work}
Recently, correlation filtering \cite{Bolme2010} has received much attention in visual tracking. It achieves state-of-the-art tracking performance in terms of both accuracy and running speed. By exploiting the circulant structure \cite{Henriques2012}, visual tracking can be described as a correlation filtering problem, which is also demonstrated to be equivalent to a ridge regression problem \cite{Henriques2012}. In this paradigm, the cyclic shifts of the latest target region (a base image) is utilized to generate a large number of training samples, essentially as the dense sampling method does, as illustrated in Fig. \ref{fig:cyclic}. The cyclic shifts lead to the fact that the sample matrix has a circulant structure. To efficient solve the correlation filtering, the tracking is conducted in frequency domain by fast Fourier transform (FFT) under the Parseval's identity. Because the filter response is considered to be of Gaussian shaped, there is a closed-form solution to the problem of the correlation filter learning. This is why the correlation filtering based tracking methods significantly improve the tracking speed.

There is extensive literature on correlation filtering based tracking methods in recent years. Henriques \etal \cite{Henriques2015} proposed to incorporating the kernel trick with the correlation filter learning, leading to kernelized version of \cite{Henriques2012}. Since the scale variations of the target appearance between successive frames are not considered in \cite{Henriques2015}, Danelljan \etal \cite{Danelljan2014b} and Li and Zhu \cite{Li2014a} integrated adaptive scale estimations with the correlation filter learning, respectively. An approach leveraging adaptive color attributes \cite{Danelljan2014a} was proposed for real-time visual tracking within the correlation filtering framework. Ma \etal \cite{Ma2015} developed a long-term correlation tracking method by decomposing visual tracking into translation and scale estimations. Liu \etal \cite{Liu2015} designed an adaptive correlation filter to exploit the part-based information of the target. Tang and Feng \cite{Tang2015} proposed a multi-kernel correlation filter for visual tracking, which fully takes advantage of the invariance-discriminative power spectrums of various features. Danelljan \etal \cite{Danelljan2015} leveraged a spatial regularization for correlation filter learning, leading to an impressive tracking performance.

Beyond the correlation filter based method, extensive tracking approaches were proposed and achieved state-of-the-art performance, such as structural learning \cite{Hare2011,Kalal2012,Zhang2015a}, sparse and low-rank learning \cite{Mei2011,Zhang2012b,Sui2015c,Sui2016}, subspace learning \cite{Kwon2010,Wang2013,Sui2015b}, and deep learning \cite{Huang2015,Wang2015}. Readers are recommended to refer to \cite{Yilmaz2006,Smeulders2014} for a thorough review of visual tracking.

\section{Proposed Approach}
\subsection{Formulation}
The typical correlation filtering based model focuses on solving the following ridge regression problem
\begin{equation}
\label{eq:ridge_regression}
\min_\w\sum_i\left(f\left(\x_i\right)-y_i\right)^2+\lambda\left\|\w\right\|_2^2,
\end{equation}
where a regression function $f\left(\x_i\right)=\w^T\varphi\left(\x_i\right)$ is trained with a feature-space projector $\varphi\left(\cdot\right)$; the objective values $y_i$ are specified to be of Gaussian shaped; and $\lambda>0$ is a weight parameter. The training samples $\left\{\x_i\right\}$ consists of the cyclically shifted image patches of the base image (\ie, the latest target). With the learned regression model, the target is localized by selecting the candidate with the largest regression value (filter response in the frequency domain) from a set of target candidates that are generated by the cyclically shifted patches of the latest target region in the current frame.

The goal of the proposed approach is to promote the robustness of the correlation filter learning. An anisotropy of the filter response, from a signal processing perspective, is exploited for visual tracking, and the robust loss functions, from a overfitting point of view, are utilized to deal with the significant appearance changes. To this end, an adaptive approach is leveraged in this work, which employs different sparsity related loss functions to adaptively fit the Gaussian shaped objective values. Similar to the previous work \cite{Henriques2012,Henriques2015}, the proposed approach is modeled from the regression perspective and solved via the correlation filtering method. Generally, the regression in this work is defined as
\begin{equation}
\label{eq:regression}
\min_\w\sum_i\ell\left(f\left(\x_i\right)-y_i\right)+\lambda\left\|\w\right\|_2^2,
\end{equation}
where $\ell\left(\cdot\right)$ is a loss function, and the regularization $\left\|\w\right\|_2^2$ is reserved to make the regression stable. In order to promote the robustness of the above model against the significant target appearance changes, the sparsity related loss function \cite{Wright2010} is encouraged. Three loss functions, $\ell_1$-, $\ell_1\ell_2$- and $\ell_{2,1}$-loss, are leveraged in this work, which exploit the sparsity, elastic net and group sparsity structures of the loss values. Note that the problem in Eq. \eqref{eq:ridge_regression} is also described via the above model when the loss function is set as $\ell_2$-loss.

\subsection{Evaluation Algorithm}
The problem in Eq. \eqref{eq:regression} is NP-hard \cite{Wright2010} because the sparsity related constraints on the data fitting term are involved. For this reason, it is equivalently reformulated as
\begin{equation}
\label{eq:regression_1}
\min_{\w,\e}\sum_i\ell\left(e_i\right)+\lambda\left\|\w\right\|_2^2,~~s.t.~e_i=y_i-f\left(\x_i\right),
\end{equation}
where $e_i$ denotes the difference between the regression values $f\left(\x_i\right)$ and the objective values $y_i$, and $y_i$ is of Gaussian shaped. Notice that the reformulated problem is convex with respect to either $\w$ or $\e$. However, it is still NP-hard with respect to both $\w$ and $\e$. As a result, an iterative algorithm is required to approximate the solution. First, an equivalent form is employed to represent the above problem as
\begin{equation}
\label{eq:ours}
\min_{\w,\e}\sum_i\left(f\left(\x_i\right)+e_i-y_i\right)^2+\lambda\left\|\w\right\|_2^2+\tau\sum_i\ell\left(e_i\right),
\end{equation}
where $\tau$ is a weight parameter. Note that Eq. \eqref{eq:ours} can be split into two subproblems:
\begin{equation}
\label{eq:w}
\min_{\w}\left\|\f\left(\X\right)+\e-\y\right\|_2^2+\lambda\left\|\w\right\|_2^2
\end{equation}
\begin{equation}
\label{eq:e}
\min_{\e}\left\|\f\left(\X\right)+\e-\y\right\|_2^2+\tau\ell\left(\e\right),
\end{equation}
where $\X$ denotes the sample matrix, of which each row denotes a sample. Both the above two subproblems have globally optimal solutions. The problem in Eq. \eqref{eq:ours} can be solved by alternately optimizing the two subproblems until the objective function values converged.

The dual space is leveraged to solve Eq. \eqref{eq:w}. The dual conjugate of $\w$, denoted by $\aa$, is introduced, such that $\w=\sum_i\alpha_i\varphi\left(\x_i\right)$. The problem with respect to $\aa$ is squared. It indicates that there is a closed-form solution
\begin{equation}
\label{eq:ouraf}
\hat{\aa}=\frac{\hat{\y}-\hat{\e}}{\hat{\k}_1+\lambda},
\end{equation}
where $\k_1$ denotes the first row of the kernel matrix $\K$ whose element $k_{ij}=\varphi^T\left(\x_i\right)\varphi\left(\x_j\right)$, the fraction means element-wise division, and the hat $\hat{~}$ stands for the discrete Fourier transform (DFT) and hereafter. Note that because the sample matrix $\X$ denotes all the training samples that are generated by cyclically shifting the latest target, some kernels, such as Gaussian, and polynomial, can lead to a circulant kernel matrix, as demonstrated in \cite{Henriques2015}. Based on such a circulant structure, the kernel matrix $\K$ can be diagonalized as
\begin{equation}
\K=\D diag\left(\hat{\k}_1\right)\D^H,
\end{equation}
where $\D$ denotes the DFT matrix, $\k_1$ denotes the first row\footnote{The rows of the kernel matrix $\K$ are actually obtained from the cyclic shifts of the vector $\k_1$.} of the kernel matrix $\K$, and $\D^H$ denotes the Hermitian transpose of $\D$. Note that the above diagonalization significantly improves the computational efficiency.

Three algorithms are employed to solve Eq. \eqref{eq:e}, corresponding to the three loss functions used in Eq. \eqref{eq:ours}.

1) \emph{$\ell_1$-loss}. In this case, the sparsity constraint is imposed on $\e$. By using the shrinkage thresholding algorithm \cite{Beck2009}, the globally optimal solution of $\e$ can be obtained from
\begin{equation}
\label{eq:e1}
\e=\sigma\left(\frac{1}{2}\tau,\F^{-1}\left(\hat{\y}-\hat{\aa}\odot\hat{\k}_1\right)\right),
\end{equation}
where $\F^{-1}\left(\cdot\right)$ denotes the inverse Fourier transform, and $\odot$ denotes the element-wise multiplication, and the function $\sigma$ is a shrinkage operator, defined as
\begin{equation}
\label{eq:shrinkage}
\sigma\left(\varepsilon,x\right)=sign\left(x\right)\max\left(0,\left|x\right|-\varepsilon\right).
\end{equation}

2) \emph{$\ell_1\ell_2$-loss}. In this case, the elastic net constraint is enforced on $\e$. By completing the square, Eq. \eqref{eq:e} can be solved in a similar way as using $\ell_1$-loss. The globally optimal solution of $\e$ is obtained from
\begin{equation}
\label{eq:e12}
\e=\sigma\left(\frac{\tau}{4+2\tau},\frac{2}{2+\tau}\F^{-1}\left(\hat{\y}-\hat{\aa}\odot\hat{\k}_1\right)\right).
\end{equation}
The coefficients of the $\ell_1$- and $\ell_2$-regularization terms in the elastic net constraint are set to be equal in the experiments.

3) \emph{$\ell_{2,1}$-loss}. In this case, the variables are considered to be two-dimensional (\ie, matrix variables). Under the $\ell_{2,1}$-loss, the group sparsity of $\e$ is exploited. By using the accelerated proximal gradient method \cite{Bach2011}, the globally optimal solution of $\e$ is obtained from
\begin{equation}
\label{eq:e21}
\e_j=
\begin{cases}
\begin{array}{cc}
\left(1-\frac{1}{\tau\left\|\q_j\right\|_2}\right)\q_j, & \frac{1}{\tau}<\left\|\q_j\right\|_2 \\
\mathbf{0}, & otherwise,
\end{array}
\end{cases}
\end{equation}
where $\e_j$ denotes the $j$-th column of the matrix $\e$, and $\q=\F^{-1}\left(\hat{\y}-\hat{\aa}\odot\hat{\k}_1\right)$. In addition, considering the symmetry of the matrix $\e$, the $j$-th row of $\e$ is also zeroed for all $j\in\left\{k|\e_k=\mathbf{0}\right\}$.

The computational cost in each iteration comes from the Fourier and the inverse Fourier transforms of $\e$, which yield $\mathcal{O}\left(n\log n\right)$ complexity. The empirical results in this work show that the algorithm converges within tens of iterations. Thus, the efficiency of the proposed approach can be satisfied for a real-time tracker.

\subsection{Target Localization}
In each frame, a large number of training samples are generated by cyclically shifting the latest target region (a base image), essentially as the dense sampling method does. Given a target candidate $\x'$, the regression value of this candidate is computed in the frequency domain from
\begin{equation}
\label{eq:cf}
\hat{\f}\left(\x'\right)=\hat{\k}'\odot\hat{\aa},
\end{equation}
where $\hat{\k}'=\varphi^T\left(\x\right)\varphi\left(\x'\right)$ denotes the kernel correlation of the latest target region $\x$ and the candidate region $\x'$. The candidate with the largest regression value (filter response) $f$ is determined as the current target. Note that the above operation in Eq. \eqref{eq:cf} is actually a spatial correlation filtering over $\k'$ using the filter $\aa$ in frequency domain, because the frequency representation can lead to significant improvement in the running speed.

\subsection{Explanation on the Loss Functions}
The different sparsity related loss functions are leveraged in this work, in order to promote the robustness of the filter learning. The $\ell_1$-loss allows the errors $\e$ to be extremely large but sparse, such that the learned filter $\aa$ may ignore the significant appearance changes of the target. The $\ell_1\ell_2$-loss appends an additional $\ell_2$-loss to the $\ell_1$-loss. Note that because the $\ell_2$-loss always leads to small and dense errors, the globally uniform appearance changes, \eg, in the case of illumination variation, can be dealt with effectively. For this reason, the $\ell_1\ell_2$-loss allows for both the abrupt and the slow appearance changes. The $\ell_{2,1}$-loss exploits the relationship between the errors, such that the appearance changes in local patches can be well handled.

\begin{figure*}[t]
  \centering
  \subfigure[]{
  \includegraphics[width=0.2\linewidth]{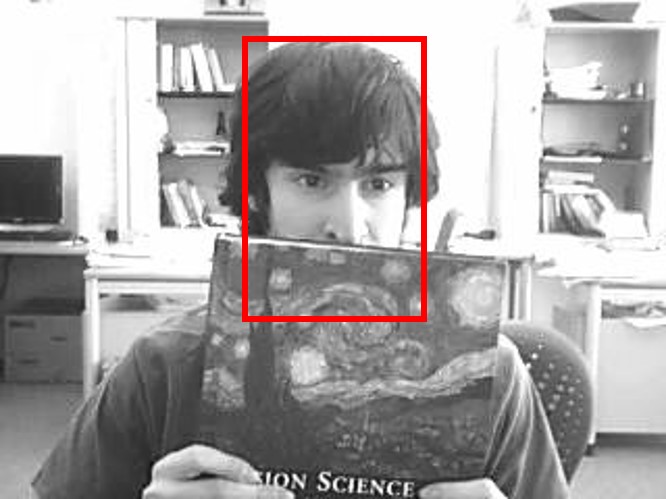}
  }\hfil
  \subfigure[]{
  \includegraphics[width=0.2\linewidth]{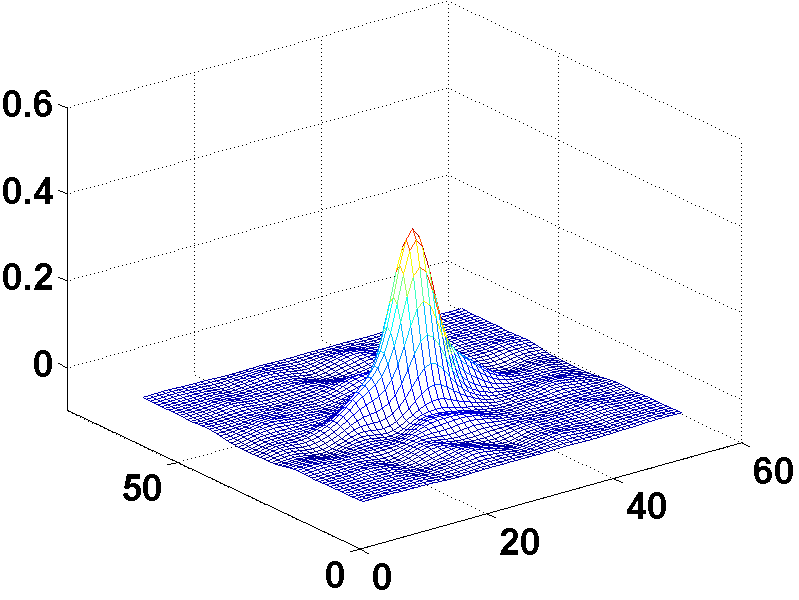}
  }\hfil
  \subfigure[]{
  \includegraphics[width=0.2\linewidth]{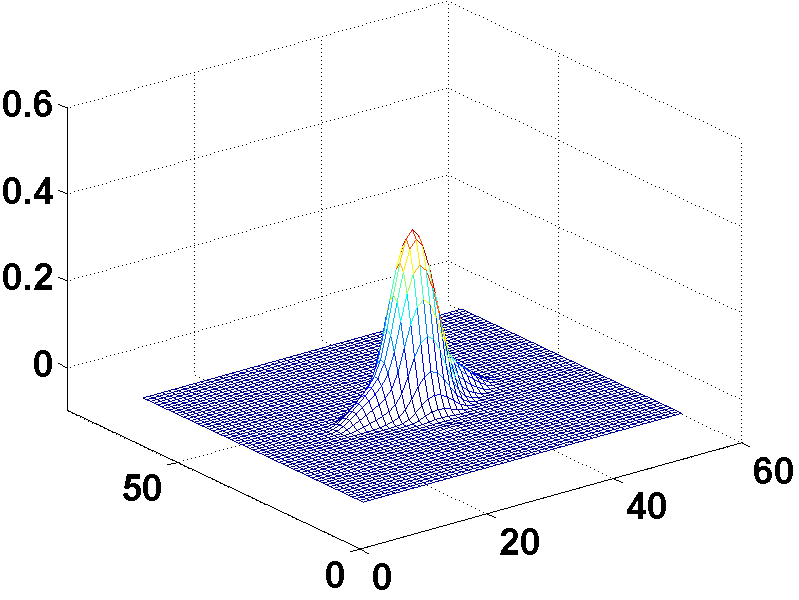}
  }\hfil
  \subfigure[]{
  \includegraphics[width=0.2\linewidth]{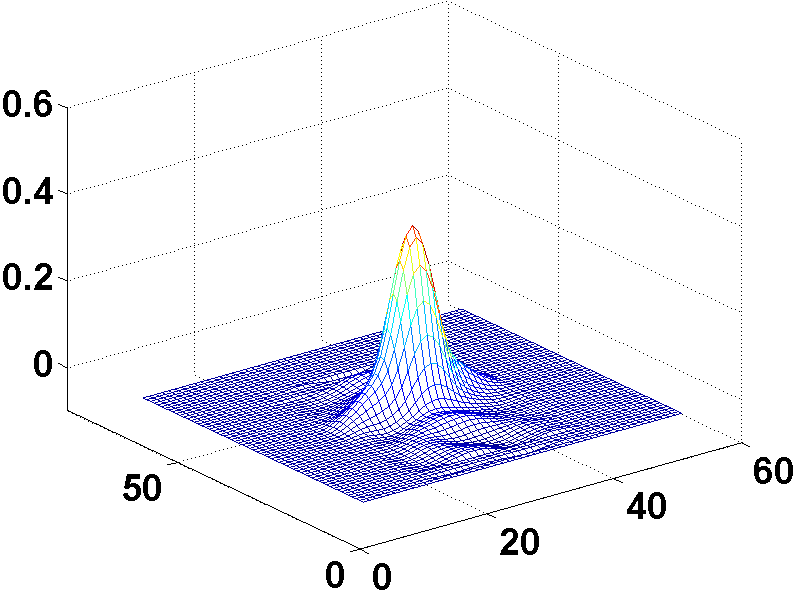}
  }
  \caption{The anisotropy of the expected filter response exploited in the frame shown in (a) with respect to the $\ell_1$-loss (b), the $\ell_1\ell_2$-loss (c), and the $\ell_{2,1}$-loss (d).}
  \label{fig:t_response}
\end{figure*}
As discussed above, the three loss functions can tolerate the large errors during the filter learning, leading to promoted robustness. Referring to Eq. \eqref{eq:regression_1}, it indicates that the difference $e_i$ between the filter response $f\left(\x_i\right)$ and the Gaussian shaped response $y_i$ can be large, leading to an anisotropic expected response $y_i-e_i$. In fact, such an anisotropy essentially facilitates tracking. Fig. \ref{fig:t_response} illustrates the anisotropic expected filter response adaptively learned via the three loss functions in a representative frame. It is evident that the three loss functions result in relatively larger filter responses in the horizontal direction. It suggests that because the distractive object (the book) moves vertically, the loss functions punish the regions vertically deviating away from the target region more severely.

\subsection{Implementation Details}
The training samples $\X$ in each frame are the fully cyclic shifts of the image region centered at the current target region with the size of 1.5 times of the target. A cosine window is leveraged in the based image to avoid the discontinuity caused by the cyclic shifts. Histogram of orientation gradient (HOG) feature is employed to describe the samples. Gaussian kernel is adopted to map the samples into a non-linear high-dimensional feature space. The above operations are also imposed on the target candidates in each frame, which cyclically shifted from the image centered at the latest target region. As recommended in \cite{Henriques2015}, the parameter $\lambda$ in Eq. \eqref{eq:ours} is set to $10^{-4}$. Another parameter $\tau$ in Eq. \eqref{eq:ours} is set to be equal to $\lambda$ in the experiments.

\section{Experiments}
Three trackers are implemented, corresponding to the $\ell_1$-, $\ell_1\ell_2$- and $\ell_{2,1}$-loss functions, denoted by Ours$_\textrm{S}$ (sparsity), Ours$_\textrm{EN}$ (elastic net), and Ours$_\textrm{GS}$ (group sparsity), respectively. The proposed trackers were evaluated on a popular benchmark \cite{Wu2013}, which contains 51 video sequences with various challenging situations, such as illumination change, non-rigid deformation, and occlusion. The target region in each frame of the 51 video sequences is labeled manually and used as the ground truth. Although many real-time trackers \cite{Grabner2006a,Zhang2013e,Zhang2012,Li2011,Hall2014,Wu2009,Bao2012,Holzer2012,Hager1996} have been proposed recently, the 12 most related state-of-the-art trackers, which are publicly provided by the authors, were compared in the experiments. Two criteria of performance evaluation were used in the comparisons, which are defined as follows.
\begin{itemize}
  \item \emph{Precision}. The percentage of frames where the center location errors (CLE) are less than a predefined threshold. The CLE in each frame is measured by the Euclidean distance between the centers of the tracking and the ground truth regions.
  \item \emph{Success Rate}. The percentage of frames where the overlap rates (OR) are greater than predefined threshold. The OR in each frame is computed from $\frac{A_t\bigcap A_g}{A_t\bigcup A_g}$ for $A_t$ and $A_g$ are the areas of the tracking and the ground truth regions, respectively.
\end{itemize}

\subsection{Comparison with the State-of-the-Art Trackers}
\begin{figure}[t]
  \centering
  \includegraphics[width=0.31\linewidth]{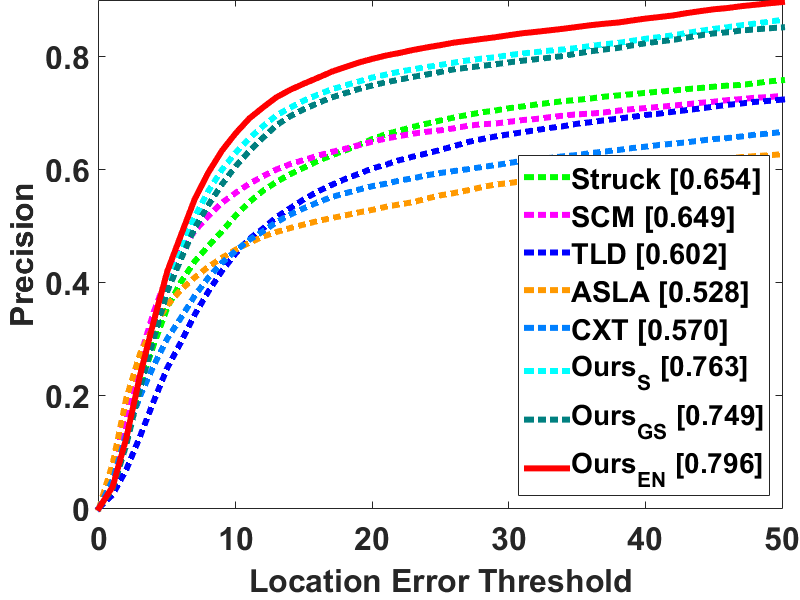}
  \hfil
  \includegraphics[width=0.31\linewidth]{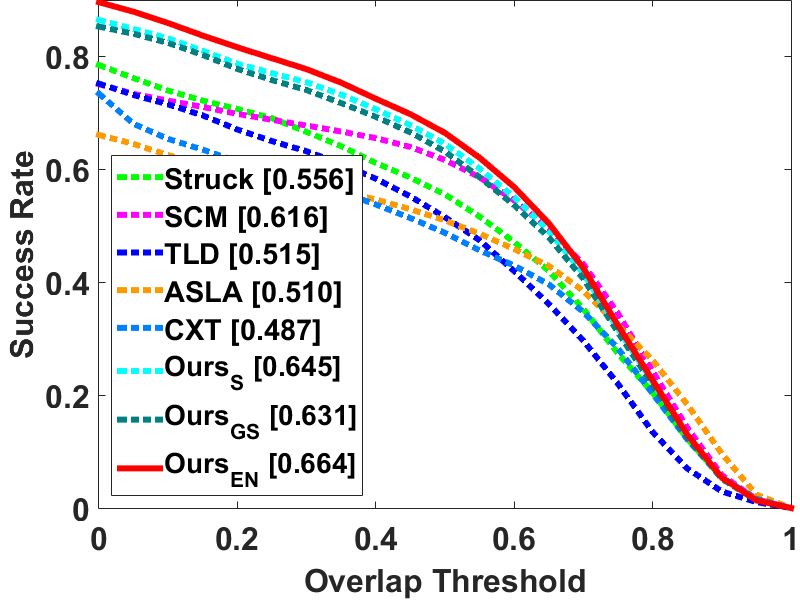}
  \caption{Tracking performance of the proposed and the top ten trackers in \cite{Wu2013} on the 51 video sequences.}
  \label{fig:top10}
\end{figure}

We compared the proposed trackers to the top five trackers in \cite{Wu2013}, including Struck \cite{Hare2011}, SCM \cite{Zhong2012}, TLD \cite{Kalal2010}, ASLA \cite{Jia2012}, and CXT \cite{Dinh2011}. Fig. \ref{fig:top10} shows the precision plots and success rate plots of the proposed and the top five trackers in \cite{Wu2013} on the 51 video sequences. It is evident that the proposed trackers significantly outperform the top five trackers, yielding the improvements of $14\%$ in precision ($\rho=20$) and $5\%$ in success rate (average). This is attributed to the advantage of the correlation filtering paradigm.

\subsection{Comparison with Trackers within Correlation Filter Learning}
\begin{figure}[t]
  \centering
  \includegraphics[width=0.31\linewidth]{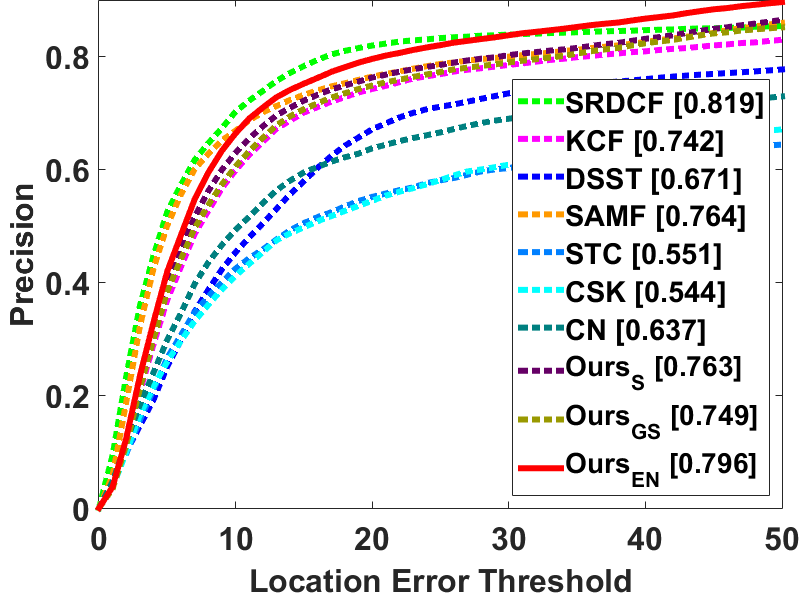}
  \hfil
  \includegraphics[width=0.31\linewidth]{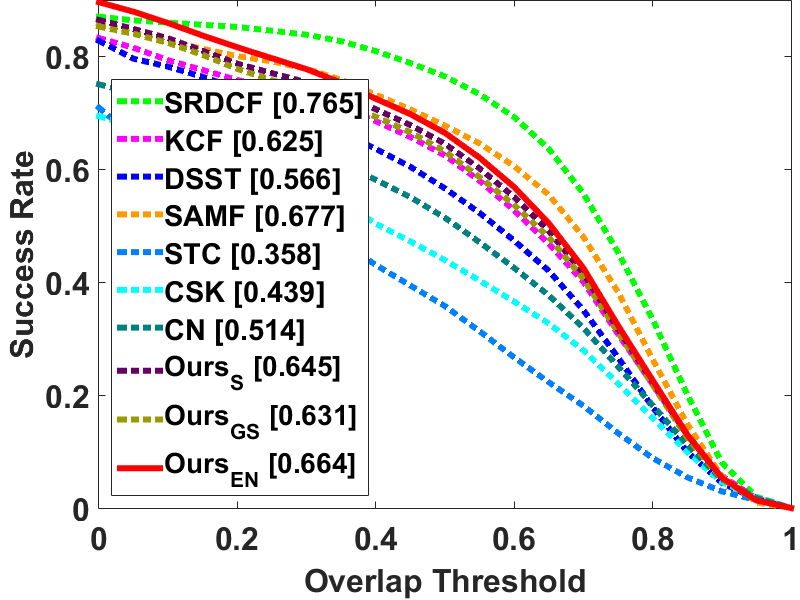}
  \caption{Tracking performance of the proposed and the popular correlation filtering based trackers on the 51 video sequences.}
  \label{fig:cf}
\end{figure}

It is also desired to investigate the performance of the proposed trackers within the correlation filtering based methods. 7 popular correlation filtering based trackers are referred to for the comparisons. Fig. \ref{fig:cf} shows the precision plots and success rate plots of the proposed and the 7 correlation filtering based trackers on the 51 video sequences.

It is evident that the proposed tracker, Ours$_\textrm{EN}$, obtains the second best results in terms of precisions. This is attributed to the robustness of the $\ell_1\ell_2$-loss used in the correlation filter learning. Because the scale variation is not considered, the performance of the proposed trackers is inferior to the SMAF tracker and the SRDCF tracker in terms of success rate, which adopt ad hoc strategies to deal with the scale change. It is also necessary to note that the success rate of the proposed trackers are still superior to other five competing counterparts (except for SMAF and SRDCF), because of the improved accuracy on the target localization.

The computational efficiency should be compared within the correlation filtering based methods for a fair evaluation. Table \ref{tab:speed} shows the running speeds (in frames per second) of the proposed and the popular correlation filtering based trackers. Because the iterative algorithm can converge within tens of iterations, the proposed trackers run faster than their counterparts like SRDCF, SMAF, and DSST.
\setlength{\tabcolsep}{2.5pt}
\begin{table}[t]
  \caption{Running speeds (in frames per second) of the proposed and the popular correlation filtering based trackers.}
  \label{tab:speed}
  \centering
  \footnotesize{
  \begin{tabular}{|l|c|c|c|c|c|c|c|c|c|c|}
    \hline
    Tracker & Ours & SRDCF \cite{Danelljan2015} & SAMF \cite{Li2014a} & DSST \cite{Danelljan2014b} & KCF \cite{Henriques2015} & CN \cite{Danelljan2014} & CSK \cite{Henriques2012} & STC \cite{Zhang2014c}\\
    \hline
    FPS & 37 & 5 & 15 & 25 & 172 & 135 & 154 & 181\\
    \hline
  \end{tabular}
  }
\end{table}

\subsection{Evaluations in Various Situations}
\begin{figure*}[t]
  \centering
  \subfigure[occlusion (31)]{
  \label{fig:occ}
  \includegraphics[width=0.225\linewidth]{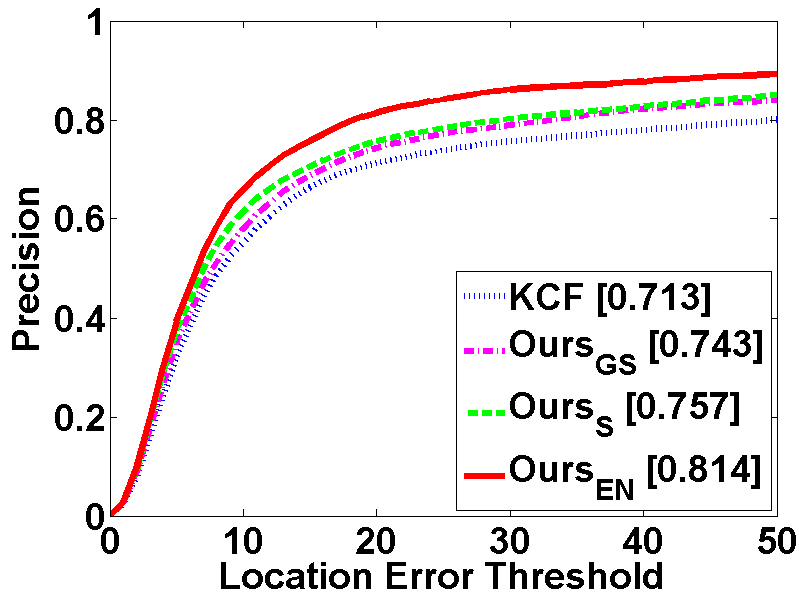}
  \includegraphics[width=0.225\linewidth]{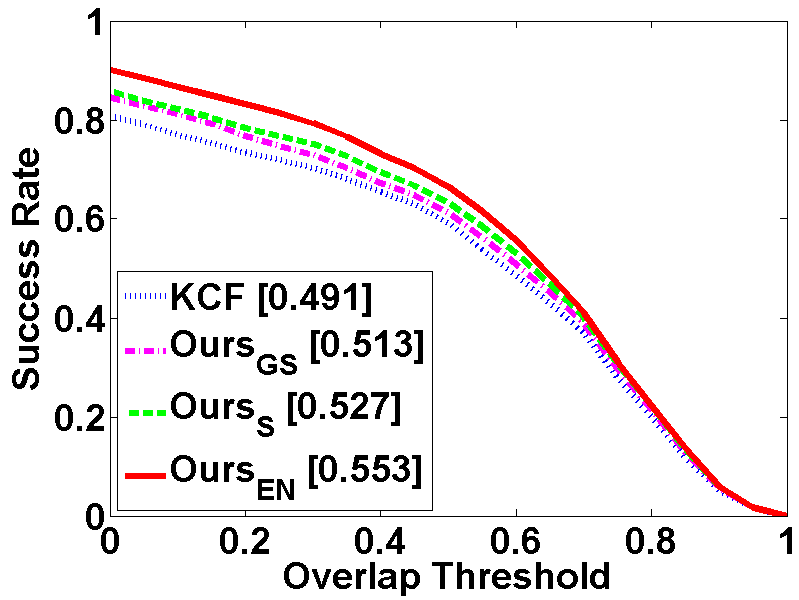}
  }
  \subfigure[deformation (19)]{
  \label{fig:def}
  \includegraphics[width=0.225\linewidth]{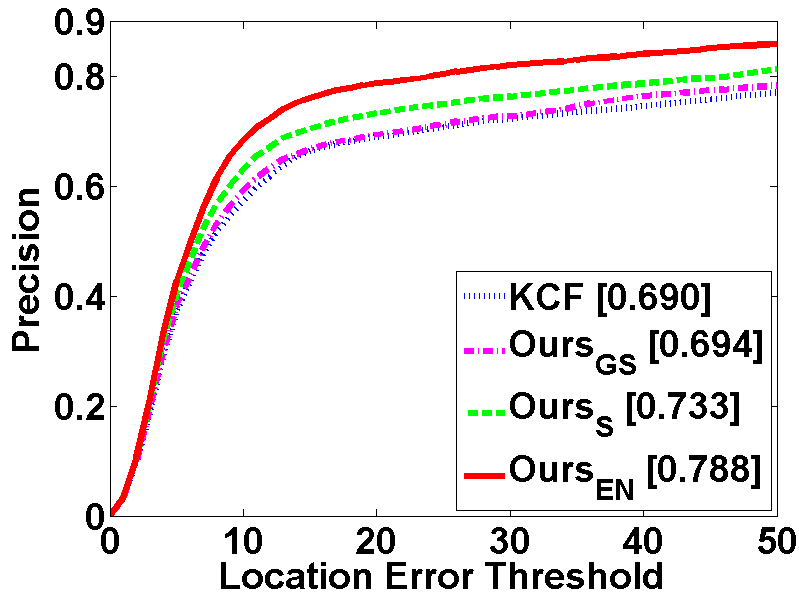}
  \includegraphics[width=0.225\linewidth]{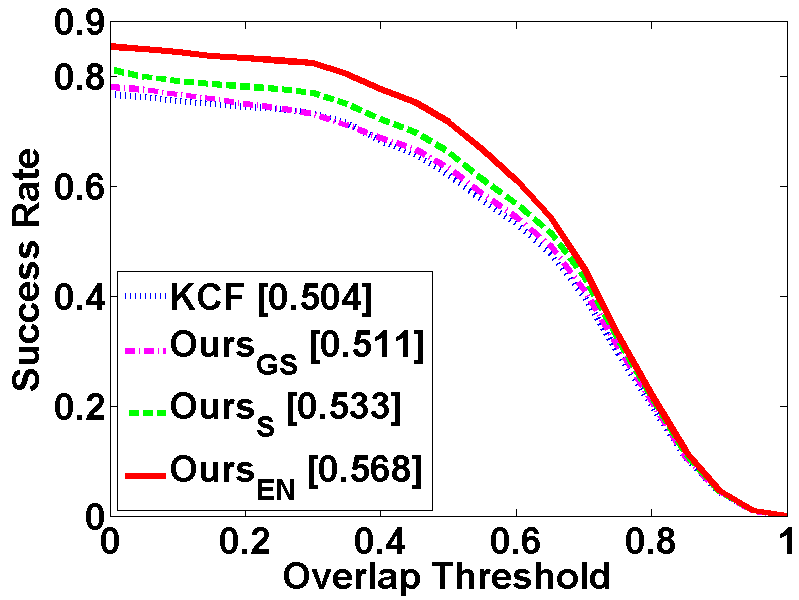}
  }
  \subfigure[out-of-plane rotation (39)]{
  \includegraphics[width=0.225\linewidth]{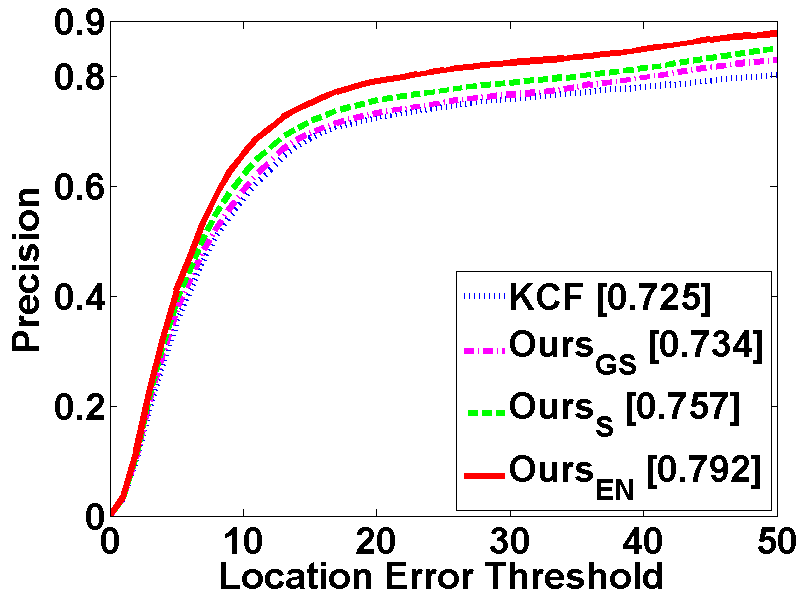}
  \includegraphics[width=0.225\linewidth]{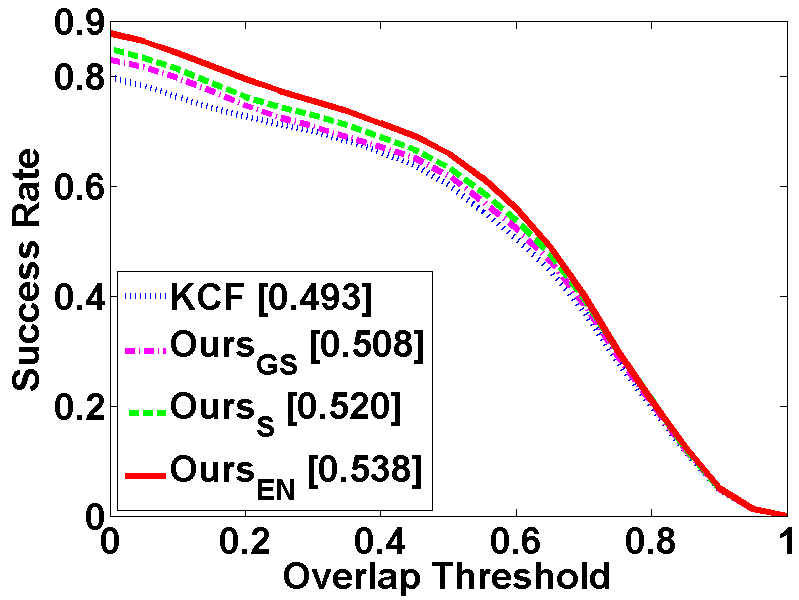}
  }
  \subfigure[in-plane rotation (31)]{
  \includegraphics[width=0.225\linewidth]{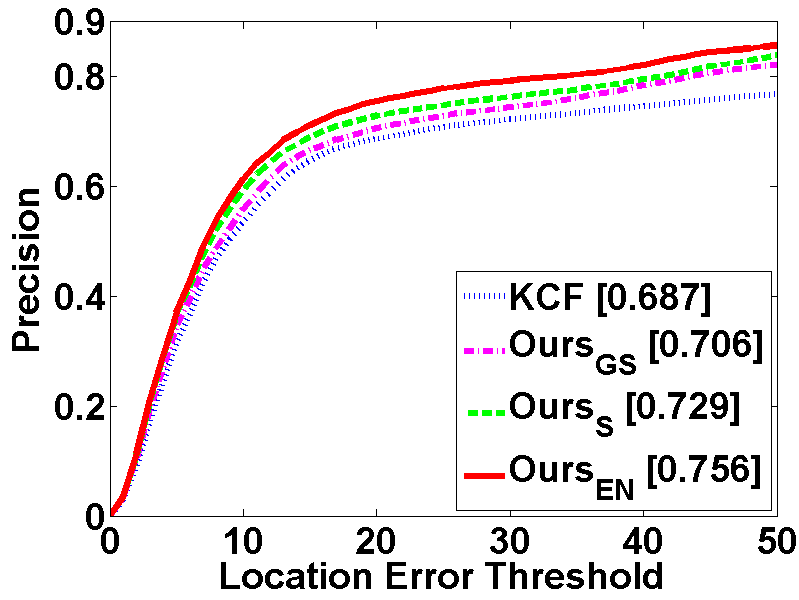}
  \includegraphics[width=0.225\linewidth]{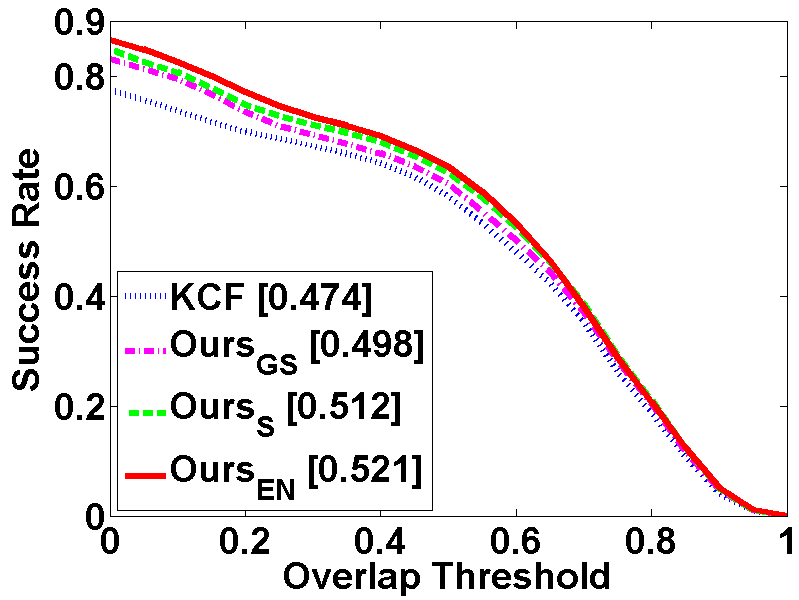}
  }
  \subfigure[illumination change (25)]{
  \label{fig:iv}
  \includegraphics[width=0.225\linewidth]{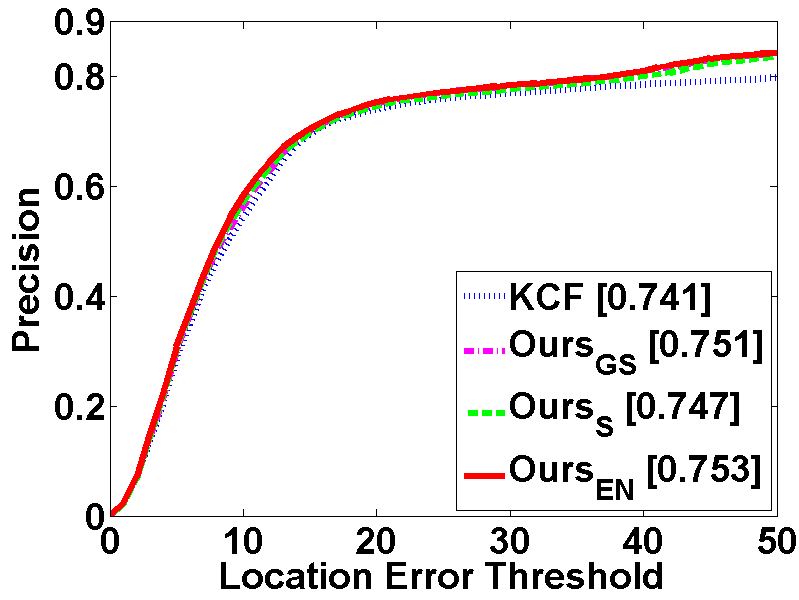}
  \includegraphics[width=0.225\linewidth]{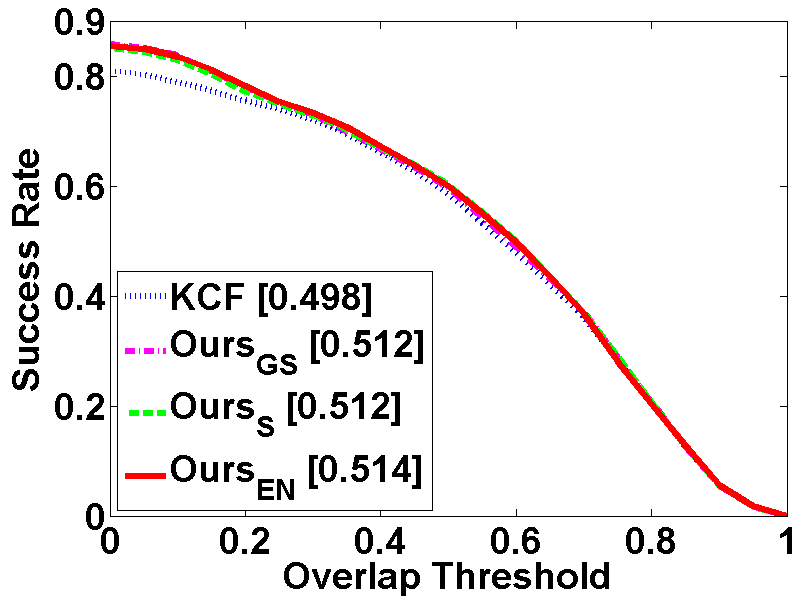}
  }
  \subfigure[scale variation (28)]{
  \includegraphics[width=0.225\linewidth]{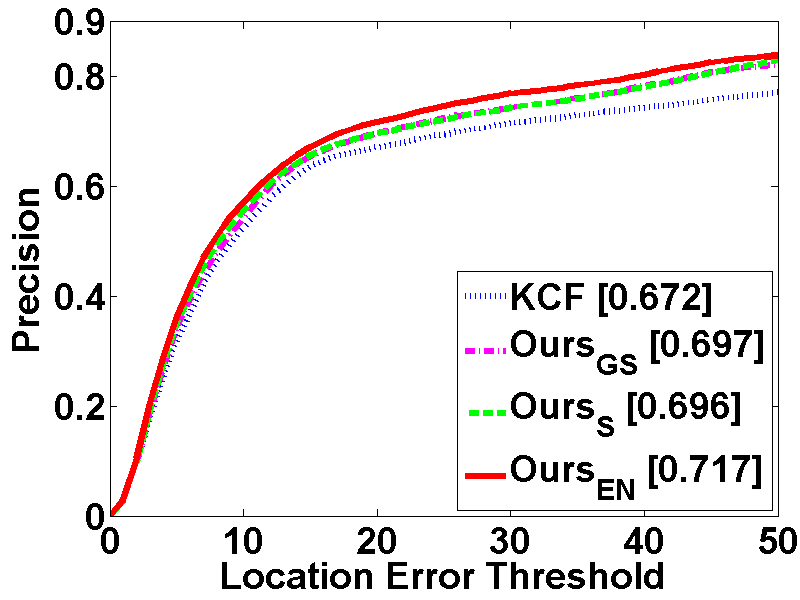}
  \includegraphics[width=0.225\linewidth]{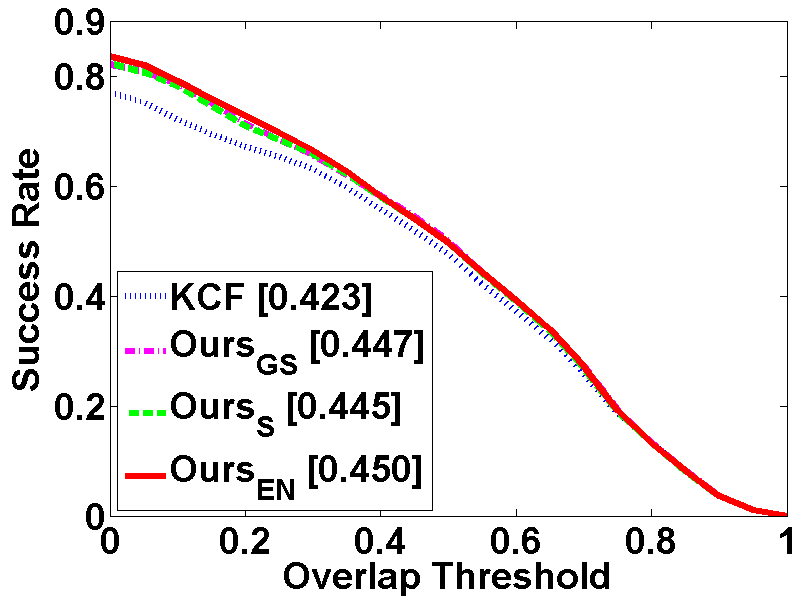}
  }
  \caption{Tracking performance of the three proposed trackers and the KCF tracker on various challenging situations. In the caption of each sub-figure, the number in parentheses denotes the number of the video sequences in the corresponding situation.}
  \label{fig:cases}
\end{figure*}

The tracking performance in various challenging situations is analyzed to thoroughly evaluate the proposed trackers. In order to investigate the effectiveness of the three ($\ell_1$-, $\ell_1\ell_2$- and $\ell_{2,1}$-) loss functions, the KCF tracker ($\ell_2$-loss) is referred to as the base line method. Fig. \ref{fig:cases} shows the results in the six challenging situations, respectively.

\emph{Occlusion}. In the case of occlusion, the target is occluded by the other objects, leading to abrupt appearance changes. It is evident that the three proposed trackers significant outperform the KCF tracker in this case. This is attributed to that the sparsity related loss functions of the proposed trackers are more robust to the abrupt appearance changes than the squared loss, resulting in more reliable filter response.

\emph{Deformation}. The target suffers from the non-rigid deformation in some complicated factors, like motion, pose change, and viewpoint variation. In this case, the target appearance often partially changes significantly. Note that the sparsity related loss functions work more robustly in the presence of significant appearance change, while the squared loss is more effective to deal with globally uniform appearance change. It is evident from the results that the proposed tracker, Ours$_\textrm{EN}$, achieves the best results in this case, because the significant changes are well handled by its sparsity constraint and the small changes are dealt with by its squared regularization. Ours$_\textrm{S}$ also obtains better results than KCF. In contrast, the $\ell_{2,1}$-loss (Ours$_\textrm{GS}$) does not improve the results obviously because of its sensitivity to this complicated situation.

\emph{In-Plane/Out-of-Plane Rotation}. This challenge is often caused by the target motion and/or viewpoint change. It is evident that the three proposed trackers improve the KCF tracker to different extent. This benefits from the robustness of the sparsity based loss functions.

\emph{Illumination Change}. In this case, the target appearance changes as the lighting condition of the scene varies. This challenge often causes uniform changes in target appearance, \ie, the illumination change influences in the entire target appearance. For this reason, the squared loss is very efficient to deal with this case. As a result, the proposed approach does not improve the KCF tracker significantly.

\emph{Scale Variation}. During tracking, the scale of the target appearance is inevitably changed. If the tracker does not adjust the size of the target window appropriately, tracking failure will be possibly caused because more background information is unexpectedly acquired by the tracker. Unfortunately, considering the efficiency, the KCF and the proposed trackers does not deal with the scales. It can be seen from the results that the proposed trackers improve the precisions but obtain similar success rates as the KCF tracker in this situation.

\subsection{Analysis of the Proposed Approach}
The goal of the proposed approach is to improve the robustness of the correlation filter learning by means of different loss functions. The different loss functions lead to different anisotropic filter responses. In this section, we interpret how the different loss functions essentially influence the tracking performance via the anisotropic filter responses.

Intuitively, the peak value of an online learned correlation filter, which is responsible for the accuracy of the target localization in each frame, should be stable enough between successive frames in the presence of various challenges. To this end, we analyze the peak values of the filter on three representative video sequences, which include the challenges of occlusion, illumination change, and deformation, respectively. Because there are also other challenges on the video sequences of \emph{faceocc2} and \emph{david}, only the first 200 and 100 frames are selected, respectively. For the convenience of discussion, we use the KCF tracker \cite{Henriques2015} as a baseline method in the analysis.

\begin{figure}[t]
  \centering
  \subfigure[]{
  \label{fig:a_occ}
  \includegraphics[width=0.48\linewidth]{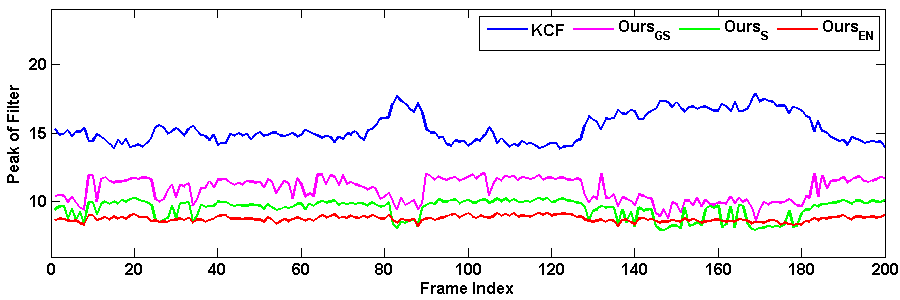}
  \includegraphics[width=0.48\linewidth]{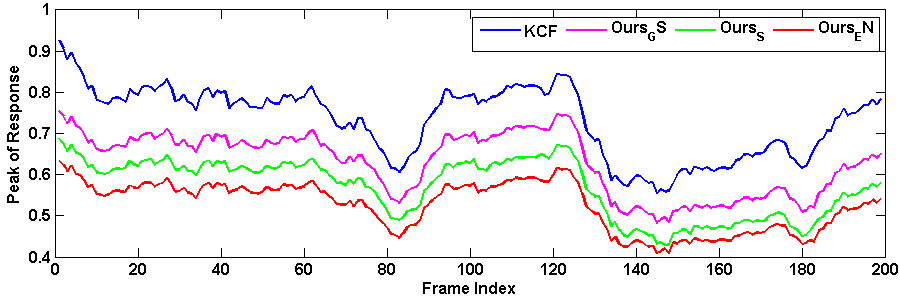}
  }
  \subfigure[]{
  \label{fig:a_iv}
  \includegraphics[width=0.48\linewidth]{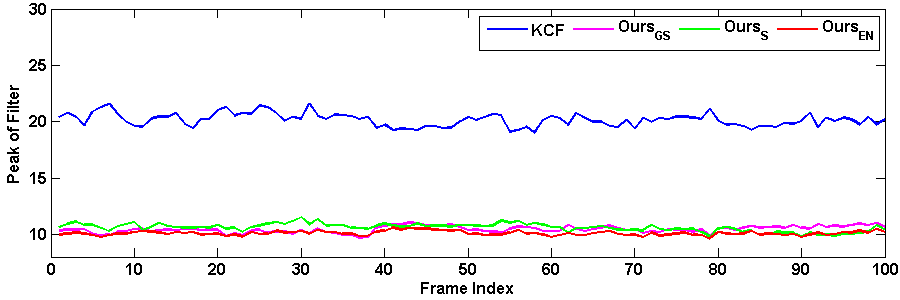}
  \includegraphics[width=0.48\linewidth]{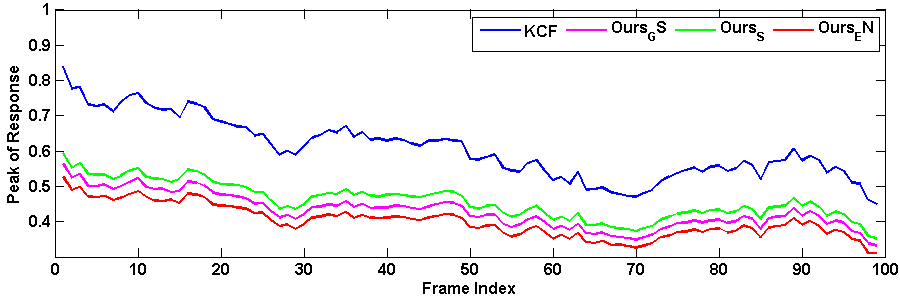}
  }
  \subfigure[]{
  \label{fig:a_def}
  \includegraphics[width=0.48\linewidth]{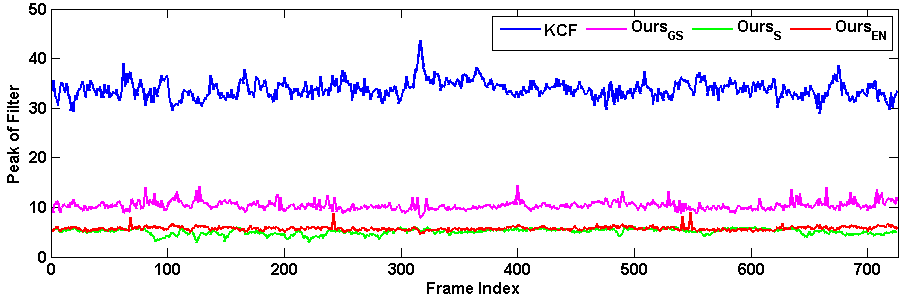}
  \includegraphics[width=0.48\linewidth]{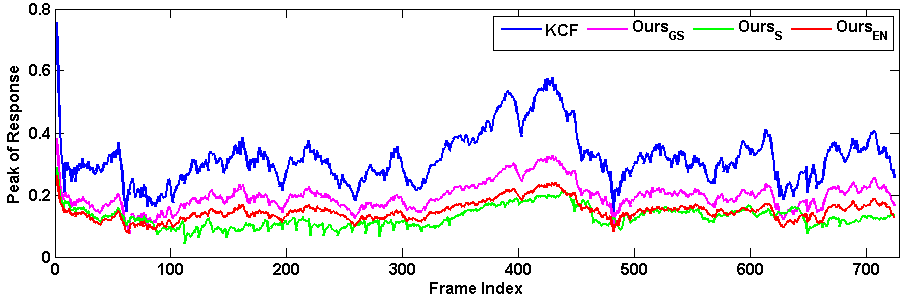}
  }
  \caption{Peak values of the online learned filters (left column) and the responses (right column) with respect to the four trackers in different challenging cases. The curves are expected to be as smooth as possible. (a) occlusion (first 200 frames of \emph{faceocc2}); (b) illumination change (first 100 frames of \emph{david}); and (c) deformation (all the 725 frames of \emph{basketball}).}
  \label{fig:a_challenge}
\end{figure}

Qualitatively, Fig. \ref{fig:a_challenge} plots the peak value of the correlation filter and the filter responses, respectively, in each frame of the three video sequences with respect to the KCF and the proposed trackers. Note that the abrupt changes of the peak values correspond to the significant appearance changes in the corresponding frames. If the peak values are sensitive in successive frames, the corresponding filter responses will be unstable, leading to lower accuracy of target location.

\begin{figure}[t]
  \centering
  \subfigure[]{
  \label{fig:ma}
  \includegraphics[width=0.31\linewidth]{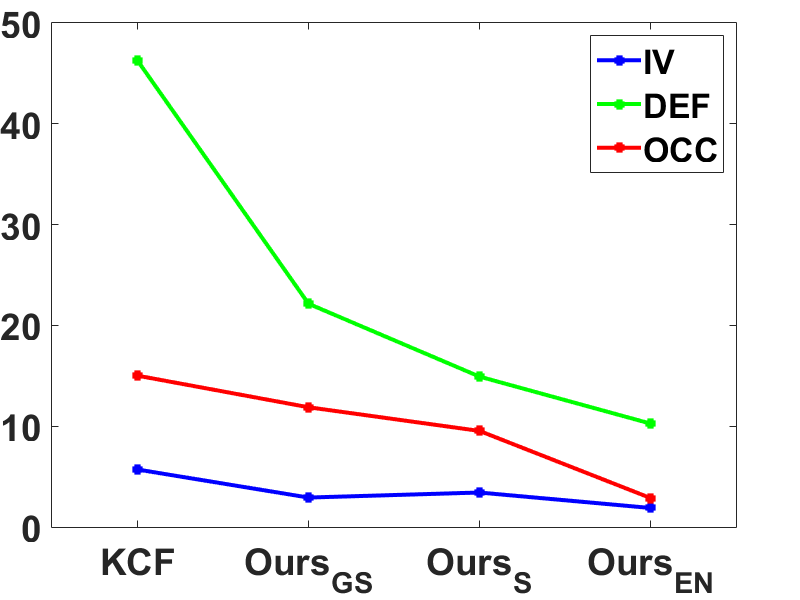}
  }
  \hfil
  \subfigure[]{
  \label{fig:mr}
  \includegraphics[width=0.31\linewidth]{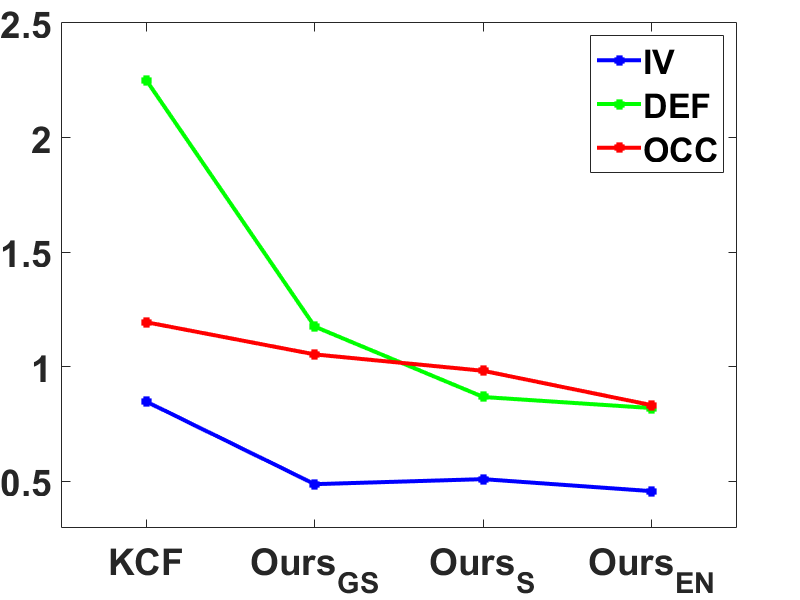}
  }
  \subfigure[]{
  \label{fig:mar}
  \includegraphics[width=0.31\linewidth]{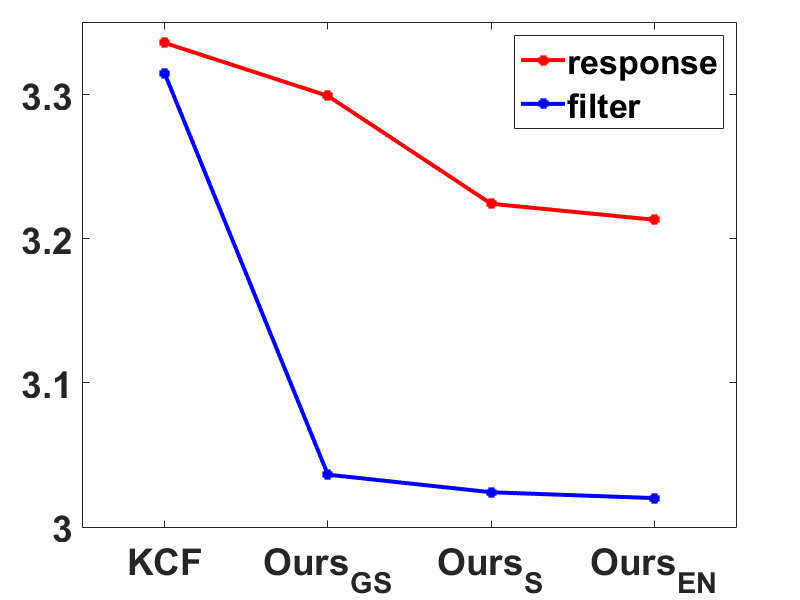}
  }
  \caption{Sensitivity of the peak values of the filters (a) and the responses (b) with respect to the four trackers in different challenging cases. (c) Average sensitivity of the peak values of the filters and the responses with respect to the four trackers on all the 51 video sequences.}
  \label{fig:m_challenge}
\end{figure}

In the case of occlusion, as shown in Fig. \ref{fig:a_occ}, the peak values of the filter with respect to the $\ell_1\ell_2$-loss (Ours$_\textrm{EN}$) varies the most slowly between the frames. The $\ell_1$-loss (Ours$_\textrm{S}$) also achieves smoother plot of the peak values than the $\ell_{2,1}$- (Ours$_\textrm{GS}$) and the $\ell_2$-loss (the KCF tracker). The analysis results on the sensitivity of the peak values in the successive frames are consistent with the tracking performance evaluations shown in Fig. \ref{fig:occ}.

In the case of illumination change, it is evident in Fig. \ref{fig:a_iv} that the peak values of the filter with respect to the four trackers have the similar sensitivities in the successive frames. It is also verified in Fig. \ref{fig:iv} that the four trackers achieve similar tracking performance.

When non-rigid deformation is involved, as shown in Fig. \ref{fig:a_def}, the proposed tracker, Ours$_\textrm{EN}$, produces the most stable filter peak values, achieving the best tracking performance, as shown in Fig. \ref{fig:def}. In contrast, the peak values of the filter with respect to the KCF tracker are very sensitive between the frames, resulting in the interior tracking performance, as evaluated in Fig. \ref{fig:def}.

Quantitatively, a metric is required to measure the sensitivity of the filter. It is discussed in \cite{Bolme2010} that, from a signal processing perspective, a good correlation filter often has a large peak-to-sidelobe ratio (PSR) value. The PSR only considers the performance of the filter in one frame, while a measurement focusing on the performance in successive frames is more desired for tracking analysis. To this end, the following metric is defined, from a visual tracking point of view, to measure the sensitivity of a correlation filter:
\begin{equation}
s=\sum_{i=1}^n{\left(p_i-p_m\right)^2},
\end{equation}
where $p_i$ denotes the peak value of the correlation filter in the $i$-th frame, $p_m$ denotes the mean of the peak values in the $n$ frames, and the $n$ peak values are normalized by their squared norm. As discussed above, the value of $s$ is expected to be small for a good correlation filter.

Figs. \ref{fig:ma} and \ref{fig:mr} plot the sensitivity $s$ of the learned correlation filters and the filter responses with respect to the four trackers in the above three challenging situations, respectively. To thoroughly verify the sensitivity, in Fig. \ref{fig:mar}, we show the average sensitivity of the filter and the response on all the 51 video sequences. It is evident that the sensitivity analysis of the correlation filter in successive frames is consistent with the tracking performance evaluations (refer to the results shown in Fig. \ref{fig:cases}).

From both the qualitative and the quantitative analysis, a conclusion can be drawn to explain how the loss functions essentially influence the tracking performance: the lower the sensitivity $s$ of the learned correlation filter in successive frames is, the higher the tracking performance is achieved. This also can be used as a criterion to design a robust correlation filter for visual tracking.

Revisiting the proposed approach, because the sparsity related loss functions allow large errors in the correlation filter learning, the appearance changes will not cause the significant changes in filter peak values, leading to low sensitivity values. In contrast, the squared loss used by the KCF tracker enforces small errors in the correlation filter learning, such that the filter is always adjusted to fit all the small appearance changes, leading to high sensitivity values. This explains why the proposed trackers perform better than the KCF tracker from the sensitivity perspective.

\subsection{Tracking in Noise Contaminated Frames}
\begin{figure}[t]
  \centering
  \includegraphics[width=0.225\linewidth]{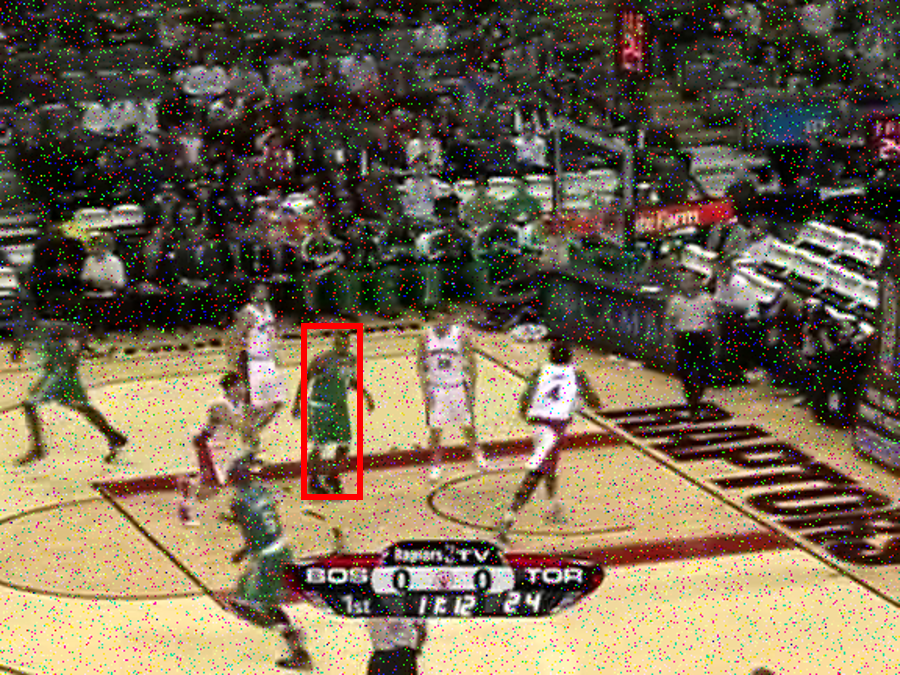}
  \includegraphics[width=0.225\linewidth]{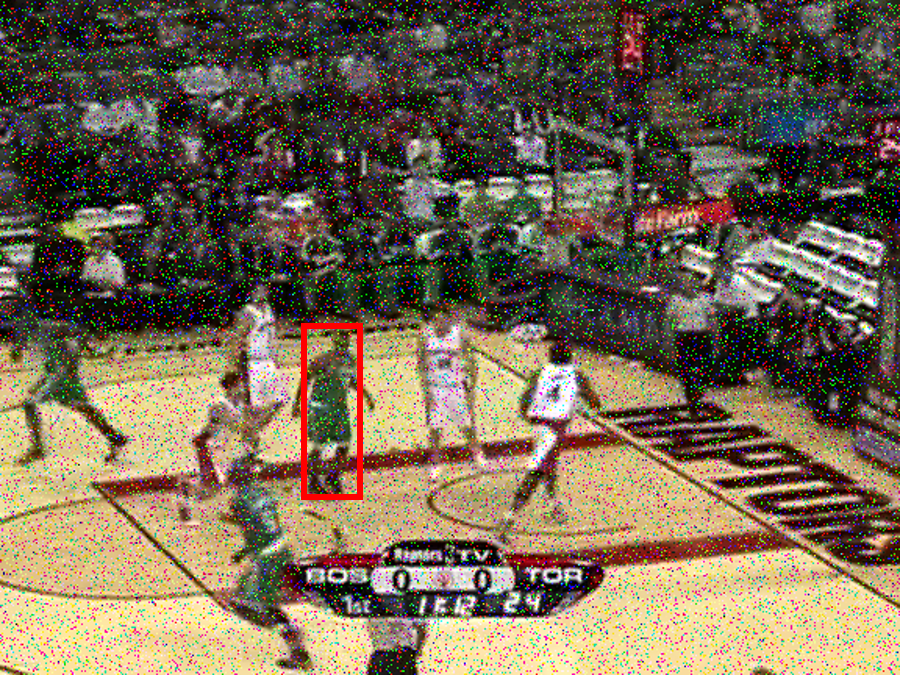}
  \includegraphics[width=0.225\linewidth]{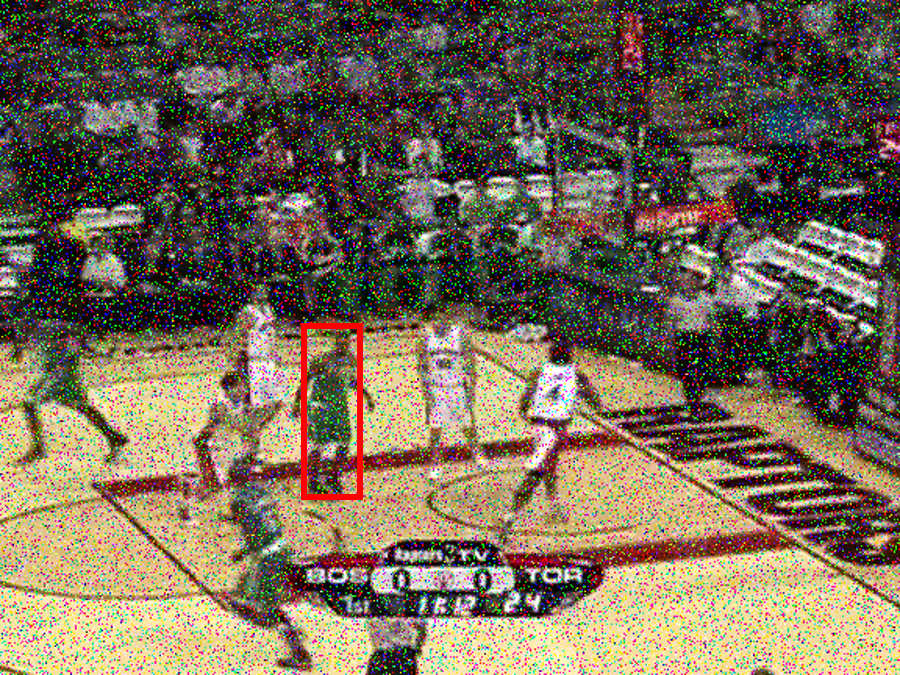}
  \includegraphics[width=0.225\linewidth]{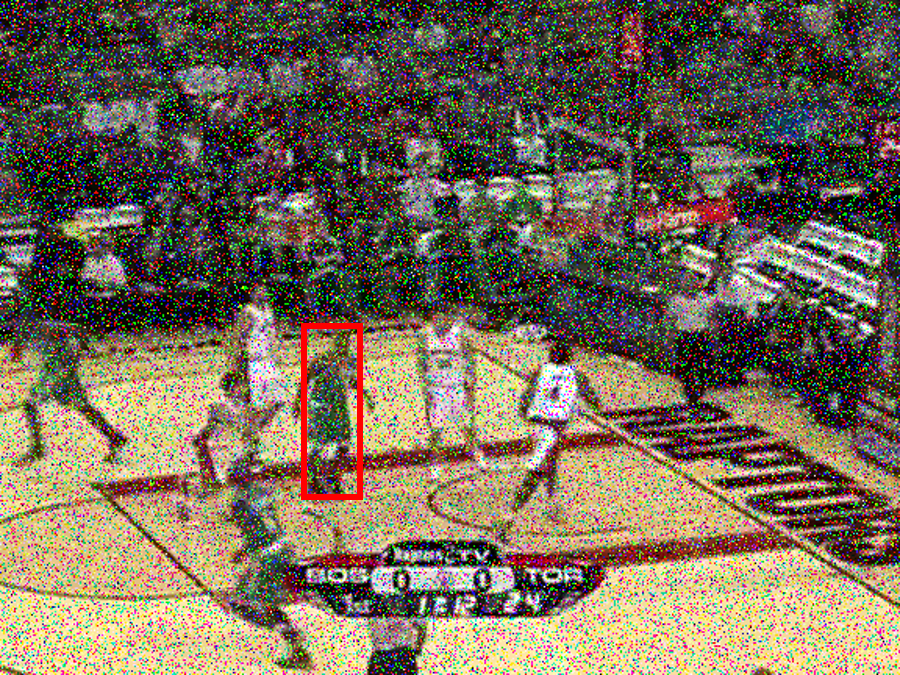}
  \caption{Representative frames with $5\%$, $10\%$ , $15\%$ and $20\%$ corrupted pixels (from left to right).}
  \label{fig:nc}
\end{figure}

\begin{figure}[t]
  \centering
  \subfigure[]{
  \includegraphics[width=0.225\linewidth]{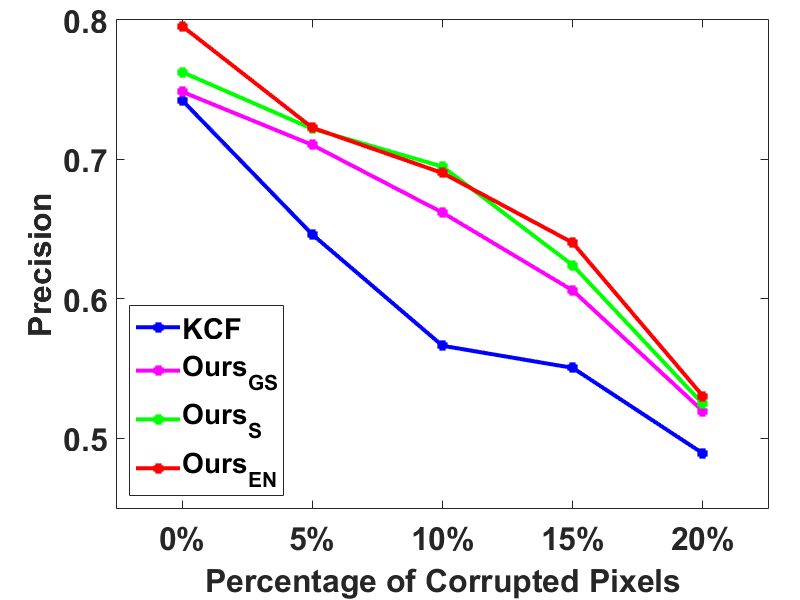}
  }
  \subfigure[]{
  \includegraphics[width=0.225\linewidth]{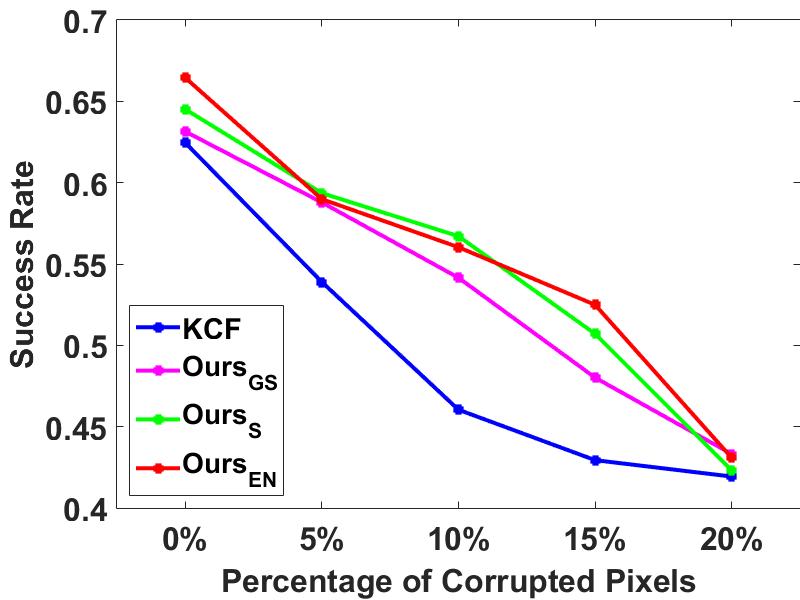}
  }
  \subfigure[]{
  \includegraphics[width=0.225\linewidth]{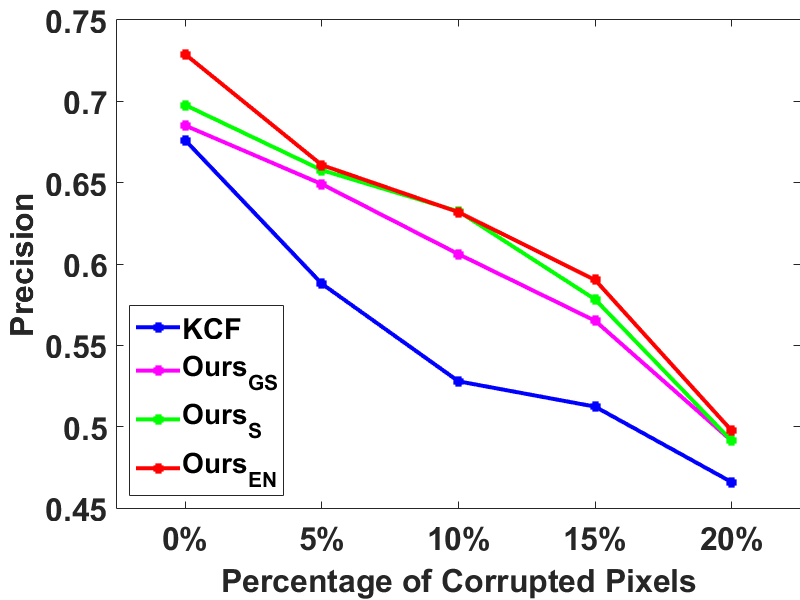}
  }
  \subfigure[]{
  \includegraphics[width=0.225\linewidth]{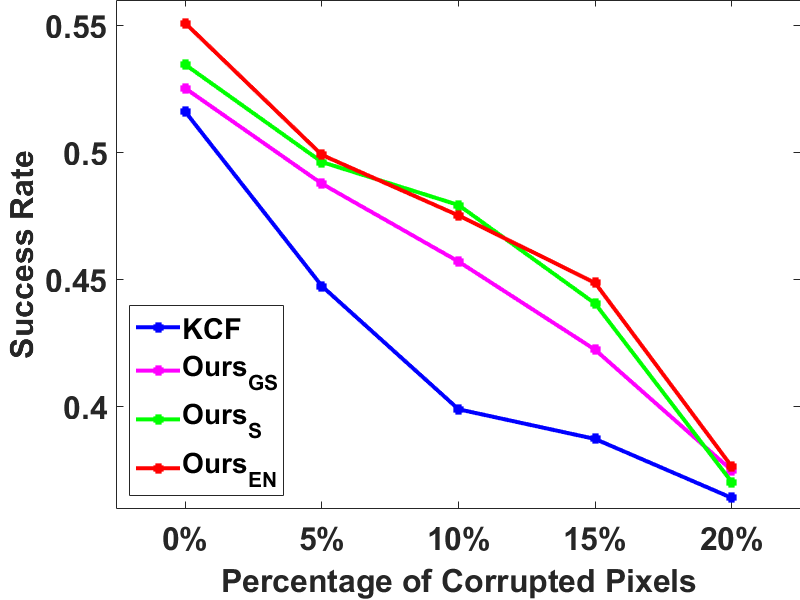}
  }
  \caption{Tracking performance of the three proposed trackers and the KCF tracker in the presence of noise with different amounts on the 51 video sequences. (a) precision plots with $\theta=20$; (b) success plot with $\rho=0.5$; (c) precision plots in average; and (d) success plot in average.}
  \label{fig:noise}
\end{figure}

In the practical applications, the quality of the video sequences cannot be guaranteed, \ie, the frames are often corrupted by noise. Thus, a visual tracker is expected to be robust to the noise contaminated frames. For this reason, to thoroughly evaluate the robustness of the proposed approach, the tracking is investigated in the noise contaminated frames. The representative noise contaminated frames are shown in Fig. \ref{fig:nc}. Fig. \ref{fig:noise} shows the tracking performance of the three proposed trackers and the KCF tracker in the presence of noise with different amounts. It is evident that, in the case that even small number of pixels are corrupted, the performance of the KCF tracker decreases significantly. In contrast, the proposed trackers are not influenced by the noise so drastically as the KCF tracker. The performance of the proposed trackers decreases sharply until a relative large number of pixels ($20\%$) are corrupted. As a result, it can be observed that the proposed trackers perform more robustly than the KCF tracker in the noise contaminated frames. This also suggests that the proposed approach is closer to the practical applications.

\section{Conclusion}
Three real-time trackers have been proposed in this work within the correlation filtering paradigm. The robustness of the filter learning has been successfully promoted by employing three sparsity related loss functions. It has been shown that the tracking performance in various challenging situations has been improved via the proposed approach. Through analyzing how the different loss functions essentially influenced the tracking performance, we have found that the analysis result on the sensitivity of the peak values of the filter is consistent with the tracking performance evaluations. This is a very useful reference criterion to design a robust correlation filter for tracking.

\subsection*{Acknowledgement}
\small{This work is partly supported by the National Natural Science Foundation of China (NSFC) under grants 61573351 and 61132007, the National Aeronautics and Space Administration (NASA) LEARN II program under grant number NNX15AN94N, and the joint fund of Civil Aviation Research by the National Natural Science Foundation of China (NSFC) and Civil Aviation Administration under grant U1533132.}


\clearpage

\begin{thebibliography}{10}
	
	\bibitem{Bolme2010}
	Bolme, D., Beveridge, J.R., Draper, B.a., Lui, Y.M.:
	\newblock {Visual object tracking using adaptive correlation filters}.
	\newblock In: IEEE Computer Society Conference on Computer Vision and Pattern
	Recognition (CVPR). (2010)
	
	\bibitem{Henriques2012}
	Henriques, F., Caseiro, R., Martins, P., Batista, J.:
	\newblock {Exploiting the Circulant Structure of Tracking-by-Detection with
		Kernels}.
	\newblock In: European Conference on Computer Vision (ECCV). (2012)  702--715
	
	\bibitem{Danelljan2014a}
	Danelljan, M., Khan, F.S., Felsberg, M., Weijer, J.V.D.:
	\newblock {Adaptive Color Attributes for Real-Time Visual Tracking}.
	\newblock In: IEEE Computer Society Conference on Computer Vision and Pattern
	Recognition (CVPR). (2014)  1090--1097
	
	\bibitem{Danelljan2014b}
	Danelljan, M., H{\"{a}}ger, G., Khan, F.S., Felsberg, M.:
	\newblock {Accurate Scale Estimation for Robust Visual Tracking}.
	\newblock In: British Machine Vision Conference (BMVC). (2014)
	
	\bibitem{Li2014a}
	Li, Y., Zhu, J.:
	\newblock {A Scale Adaptive Kernel Correlation Filter Tracker with Feature
		Integration}.
	\newblock In: European Conference on Computer Vision Workshop. (2014)
	
	\bibitem{Zhang2014c}
	Zhang, K., Zhang, L., Liu, Q., Zhang, D., Yang, M.H.:
	\newblock {Fast Visual Tracking via Dense Spatio-Temporal Context Learning}.
	\newblock In: European Conference on Computer Vision (ECCV). (2014)
	
	\bibitem{Henriques2015}
	Henriques, J., Caseiro, R., Martins, P., Batista, J.:
	\newblock {High-Speed Tracking with Kernelized Correlation Filters}.
	\newblock IEEE Transactions on Pattern Analysis and Machine Intelligence
	(TPAMI) \textbf{37}(3) (2015)  583--596
	
	\bibitem{Liu2015}
	Liu, T., Wnag, G., Yang, Q.:
	\newblock {Real-time part-based visual tracking via adaptive correlation
		filters}.
	\newblock In: IEEE Computer Society Conference on Computer Vision and Pattern
	Recognition (CVPR). (2015)  4902--4912
	
	\bibitem{Danelljan2015}
	Danelljan, M., Gustav, H., Khan, F.S., Felsberg, M.:
	\newblock {Learning Spatially Regularized Correlation Filters for Visual
		Tracking}.
	\newblock In: IEEE International Conference on Computer Vision (ICCV). (2015)
	4310--4318
	
	\bibitem{Shunli2015}
	Zhang, S., Zhao, S., Sui, Y., Zhang, L.:
	\newblock {Single object tracking with fuzzy least squares support vector
		machine}.
	\newblock IEEE Transactions on Image Processing (TIP) \textbf{24}(12) (2015)
	5723--5738
	
	\bibitem{Hare2011}
	Hare, S., Saffari, A., Torr, P.:
	\newblock {Struck: Structured output tracking with kernels}.
	\newblock In: IEEE International Conference on Computer Vision (ICCV). (2011)
	263--270
	
	\bibitem{Wu2013}
	Wu, Y., Lim, J., Yang, M.H.:
	\newblock {Online Object Tracking: A Benchmark}.
	\newblock In: IEEE Computer Society Conference on Computer Vision and Pattern
	Recognition (CVPR). (2013)  2411--2418
	
	\bibitem{Ma2015}
	Ma, C., Yang, X., Zhang, C., Yang, M.h.:
	\newblock {Long-term Correlation Tracking}.
	\newblock In: IEEE Computer Society Conference on Computer Vision and Pattern
	Recognition (CVPR). (2015)  5388--5396
	
	\bibitem{Tang2015}
	Tang, M., Feng, J.:
	\newblock {Multi-kernel Correlation Filter for Visual Tracking}.
	\newblock In: IEEE International Conference on Computer Vision (ICCV). (2015)
	3038--3046
	
	\bibitem{Kalal2012}
	Kalal, Z., Mikolajczyk, K., Matas, J.:
	\newblock {Tracking-learning-detection}.
	\newblock IEEE Transactions on Pattern Analysis and Machine Intelligence
	(TPAMI) \textbf{34}(7) (2012)  1409--1422
	
	\bibitem{Zhang2015a}
	Zhang, T., Liu, S., Xu, C., Yan, S., Ghanem, B., Ahuja, N., Yang, M.h.:
	\newblock {Structural Sparse Tracking}.
	\newblock In: IEEE Computer Society Conference on Computer Vision and Pattern
	Recognition (CVPR). (2015)  150--158
	
	\bibitem{Mei2011}
	Mei, X., Ling, H.:
	\newblock {Robust visual tracking and vehicle classification via sparse
		representation}.
	\newblock IEEE Transactions on Pattern Analysis and Machine Intelligence
	(TPAMI) \textbf{33}(11) (nov 2011)  2259--2272
	
	\bibitem{Zhang2012b}
	Zhang, T., Ghanem, B., Liu, S., Ahuja, N.:
	\newblock {Low-rank sparse learning for robust visual tracking}.
	\newblock In: European Conference on Computer Vision (ECCV). (2012)  470--484
	
	\bibitem{Sui2015c}
	Sui, Y., Tang, Y., Zhang, L.:
	\newblock {Discriminative Low-Rank Tracking}.
	\newblock In: IEEE International Conference on Computer Vision (ICCV). (2015)
	3002--3010
	
	\bibitem{Sui2016}
	Sui, Y., Zhang, L.:
	\newblock {Robust Tracking via Locally Structured Representation}.
	\newblock International Journal of Computer Vision (IJCV) \textbf{119}(2)
	(2016)  110--144
	
	\bibitem{Kwon2010}
	Kwon, J., Lee, K.:
	\newblock {Visual tracking decomposition}.
	\newblock In: IEEE Computer Society Conference on Computer Vision and Pattern
	Recognition (CVPR). (2010)  1269--1276
	
	\bibitem{Wang2013}
	Wang, D., Lu, H., Yang, M.H.:
	\newblock {Least Soft-thresold Squares Tracking}.
	\newblock In: IEEE Computer Society Conference on Computer Vision and Pattern
	Recognition (CVPR). (2013)  2371--2378
	
	\bibitem{Sui2015b}
	Sui, Y., Zhang, S., Zhang, L.:
	\newblock {Robust Visual Tracking via Sparsity-Induced Subspace Learning}.
	\newblock IEEE Transactions on Image Processing (TIP) \textbf{24}(12) (2015)
	4686--4700
	
	\bibitem{Huang2015}
	Ma, C., Huang, J.b., Yang, X., Yang, M.H.:
	\newblock {Hierarchical Convolutional Features for Visual Tracking}.
	\newblock In: IEEE International Conference on Computer Vision (ICCV). (2015)
	3074--3082
	
	\bibitem{Wang2015}
	Wang, L., Ouyang, W., Wang, X., Lu, H.:
	\newblock {Visual Tracking with Fully Convolutional Networks}.
	\newblock In: IEEE International Conference on Computer Vision (ICCV). (2015)
	3119--3127
	
	\bibitem{Yilmaz2006}
	Yilmaz, A., Javed, O., Shah, M.:
	\newblock {Object tracking: A Survey}.
	\newblock ACM Computing Surveys \textbf{38}(4) (dec 2006)  13--57
	
	\bibitem{Smeulders2014}
	Smeulders, A.W.M., Chu, D.M., Cucchiara, R., Calderara, S., Dehghan, A., Shah,
	M.:
	\newblock {Visual Tracking: An Experimental Survey}.
	\newblock IEEE Transactions on Pattern Analysis and Machine Intelligence
	(TPAMI) \textbf{36}(7) (nov 2014)  1442--1468
	
	\bibitem{Wright2010}
	Wright, J., Ma, Y., Mairal, J., Sapiro, G.:
	\newblock {Sparse representation for computer vision and pattern recognition}.
	\newblock Proceedings of The IEEE \textbf{98}(6) (2010)  1031--1044
	
	\bibitem{Beck2009}
	Beck, A., Teboulle, M.:
	\newblock {A Fast Iterative Shrinkage-Thresholding Algorithm for Linear Inverse
		Problems}.
	\newblock SIAM Journal on Imaging Sciences \textbf{2}(1) (jan 2009)  183--202
	
	\bibitem{Bach2011}
	Bach, F., Jenatton, R., Mairal, J., Obozinski, G.:
	\newblock {Convex optimization with sparsity-inducing norms}.
	\newblock Optimization for Machine Learning (2011)  1--35
	
	\bibitem{Grabner2006a}
	Grabner, H., Grabner, M., Bischof, H.:
	\newblock {Real-Time Tracking via On-line Boosting}.
	\newblock In: British Machine Vision Conference (BMVC). (2006)  6.1--6.10
	
	\bibitem{Zhang2013e}
	Zhang, K., Zhang, L., Yang, M.H.:
	\newblock {Real-Time Object Tracking via Online Discriminative Feature
		Selection}.
	\newblock IEEE Transactions on Image Processing (TIP) \textbf{22}(12) (dec
	2013)  4664--4677
	
	\bibitem{Zhang2012}
	Zhang, K., Zhang, L., Yang, M.H.:
	\newblock {Real-Time Compressive Tracking}.
	\newblock In: European Conference on Computer Vision (ECCV). (2012)  866--879
	
	\bibitem{Li2011}
	Li, H., Shen, C., Shi, Q.:
	\newblock {Real-time visual tracking using compressive sensing}.
	\newblock In: IEEE Computer Society Conference on Computer Vision and Pattern
	Recognition (CVPR). (2011)
	
	\bibitem{Hall2014}
	Hall, D., Perona, P.:
	\newblock {Online, Real-Time Tracking Using a Category-to-Individual Detector}.
	\newblock In: European Conference on Computer Vision (ECCV). (2014)  361--376
	
	\bibitem{Wu2009}
	Wu, Y., Cheng, J., Wang, J., Lu, H.:
	\newblock {Real-time visual tracking via incremental covariance tensor
		learning}.
	\newblock In: IEEE International Conference on Computer Vision (ICCV). (2009)
	
	\bibitem{Bao2012}
	Bao, C., Wu, Y., Ling, H., Ji, H.:
	\newblock {Real time robust L1 tracker using accelerated proximal gradient
		approach}.
	\newblock In: IEEE Computer Society Conference on Computer Vision and Pattern
	Recognition (CVPR). (jun 2012)  1830--1837
	
	\bibitem{Holzer2012}
	Holzer, S., Pollefeys, M., Ilic, S., Tan, D., Navab, N.:
	\newblock {Online Learning of Linear Predictors for Real-Time Tracking}.
	\newblock In: European Conference on Computer Vision (ECCV). (2012)
	
	\bibitem{Hager1996}
	Hager, G.D., Belhumeur, P.N.:
	\newblock {Real-Time Tracking of Image Regions with Changes in Geometry and
		Illumination}.
	\newblock In: IEEE Computer Society Conference on Computer Vision and Pattern
	Recognition (CVPR). (1996)  403--410
	
	\bibitem{Zhong2012}
	Zhong, W., Lu, H., Yang, M.H.:
	\newblock {Robust Object Tracking via Sparsity-based Collaborative Model}.
	\newblock In: IEEE Computer Society Conference on Computer Vision and Pattern
	Recognition (CVPR). (2012)  1838--1845
	
	\bibitem{Kalal2010}
	Kalal, Z., Matas, J., Mikolajczyk, K.:
	\newblock {P-N learning: Bootstrapping binary classifiers by structural
		constraints}.
	\newblock In: IEEE Computer Society Conference on Computer Vision and Pattern
	Recognition (CVPR). (jun 2010)  49--56
	
	\bibitem{Jia2012}
	Jia, X., Lu, H., Yang, M.H.:
	\newblock {Visual Tracking via Adaptive Structural Local Sparse Appearance
		Model}.
	\newblock In: IEEE Computer Society Conference on Computer Vision and Pattern
	Recognition (CVPR). (2012)  1822--1829
	
	\bibitem{Dinh2011}
	Dinh, T.B., Vo, N., Medioni, G.:
	\newblock {Context Tracker: Exploring Supporters and Distracters in
		Unconstrained Environments}.
	\newblock In: IEEE Computer Society Conference on Computer Vision and Pattern
	Recognition (CVPR). (jun 2011)  1177--1184
	
	\bibitem{Danelljan2014}
	Danelljan, M., Khan, F.S., Felsberg, M., Weijer, J.V.D.:
	\newblock {Adaptive Color Attributes for Real-Time Visual Tracking}.
	\newblock In: IEEE Computer Society Conference on Computer Vision and Pattern
	Recognition (CVPR). (2014)
	
\end{thebibliography}
\end{document}